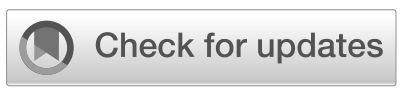

*CORRESPONDENCE
Pamela M. Burrage
pamela.burrage@qut.edu.au

# Learning surrogate equations for the analysis of an agent-based cancer model

Kevin Burrage[1,2], Pamela M. Burrage[1]*, Justin N. Kreikemeyer[3], Adelinde M. Uhrmacher[3] and Hasitha N. Weerasinghe[1]

[1]School of Mathematical Sciences, Queensland University of Technology, Brisbane, QLD, Australia,
[2]Visiting Professor at Department of Computer Science, University of Oxford, Oxford, United Kingdom,
[3]Modelling and Simulation, Institute for Visual and Analytic Computing, Faculty of Informatics and Electrical Engineering, University of Rostock, Rostock, Germany

In this paper, we adapt a two-species agent-based cancer model that describes the interaction between cancer cells and healthy cells on a uniform grid to include the interaction with a third species—namely immune cells. We run six different scenarios to explore the competition between cancer and immune cells and the initial concentration of the immune cells on cancer dynamics. We then use coupled equation learning to construct a population-based reaction model for each scenario. We show how they can be unified into a single surrogate population-based reaction model, whose underlying three coupled ordinary differential equations are much easier to analyse than the original agent-based model. As an example, by finding the single steady state of the cancer concentration, we are able to find a linear relationship between this concentration and the initial concentration of the immune cells. This then enables us to estimate suitable values for the competition and initial concentration to reduce the cancer substantially without performing additional complex and expensive simulations from an agent-based stochastic model.

## 1 Introduction

In Weerasinghe et al. [1], we developed an agent-based model to explore interactions between cancerous cells and healthy cells in the presence of proteins in the extracellular matrix (ECM). This setting is sometimes called the tumor microenvironment (TME). The model represents a two-dimensional tissue section comprising healthy cells, cancer cells and ECM proteins. The domain consists of an inner and outer region on a uniform $100 \times 100$ grid. The outer region represents the invasive margin of the tissue. The boundary of the domain represents the surrounding ECM membrane and is fully fenced with no gaps at time zero. We assume that healthy cells are located only in the inner region, and their movement is negligible as adhesion molecules keep these healthy cells in their intended locations [2]. The ECM proteins are randomly placed in the domain and do not move. Initially, the domain represents healthy tissue, and then a single healthy cell turns into a cancerous cell that spreads throughout the tissue due to a series of mutations. Cancer cells cannot occupy a grid position filled by an ECM protein.

The model records the healthy cell and cancer cell densities as a function of time, along with the number of cancer cells that exit the domain due to the breaking down of the surrounding fence. The agent-based model simulates these cell-cell and cell-ECM interactions according to a set of interaction rules underpinned by the eight hallmarks of cancer [3, 4], which allows a link between tumor heterogeneity and cellular response. The model allows a much richer set of interactions than the standard approach via unimolecular and bimolecular reactions that underpin the Law of Mass Action ordinary differential equation. These include, for example, a cell stickiness value, a jump radius, the maximum number of healthy cell divisions, a cell and division age, and a competition rate between healthy and cancer cells, to name a few—see Weerasinghe et al. [1] for more details. An example of cancer progression within the agent-based model was given in Weerasinghe [9], and we present it here—in which healthy cells, cancer cells and ECM proteins (obstacles) are represented as blue, red and black nodes (Figure 1).

Outputs from the model show that cell-cell and cell-ECM interactions affect cancer cell dynamics and that low initial healthy cell and ECM protein densities promote cancer progression, cell motility and invasion, while high ECM breakdown probabilities of cancer cells on the ECM can also promote cancer invasion.

Of course the processes underlying, for example, multiple myeloma in the bone marrow take place in an unstructured three-dimensional setting with very complex internal structures. So one of our assumptions is that our two-dimensional model taking place on a regular grid with a very simple internal structure based on inner and outer domains is still appropriate for capturing this complexity. This is of course a limitation but at the same time could be considered a benefit when using equation learning. Additionally our agent-based model is quite simple when describing the interactions between cancer cells and immune cells and the movement of cancer cells. Within the bone marrow, there is a complex set of interactions between a variety of proteins that we do not attempt to capture. This again is a limitation but it is also a strength of our approach as there is little information about the protein interaction rates and introducing complexity without such information has negative repercussions when using equation learning.

Even though reaction models following the Law of Mass Action cannot describe the full range of behaviors possible in stochastic ABMs, they form a useful abstraction that is much easier to analyse than an ABM. In this paper, we pursue two objectives. First, we extend the two-species ABM into a three-species model that incorporates treatment strategies involving immune cells (T-cells) (Section 2.1). This is designed to help with the analysis of recently developed clinical protocols that fit into the class of immunotherapy based on the injection of immune cells. We then apply equation learning techniques [5–8] to automatically construct a surrogate reaction system from six different parametrisations of the three-species ABM (Sections 2.2–2.4). These learning techniques determine ODE models from time-series measurements by sparse symbolic regression. In contrast to the stochastic ABM formulation and other kinds of (neural) surrogate models, these are interpretable by humans and can be analyzed mathematically. Finally, we show how to use this simplified representation of the ABM to determine the steady-state behavior of the model without performing costly evaluations, and to make predictions of the efficacy of treatment strategies (Section 3).

## 2 Materials and methods

### 2.1 An augmented agent-based model with immune cells

In this paper, we will adapt the two-species agent-based model in Weerasinghe et al. [1] to a three-species model in which the third species represents T-cells (immune cells) that can destroy cancer cells. The key elements of the agent-based cancer model, with regard to a two-species interaction between cancer cells and healthy cells, were described in Weerasinghe et al. [1]. In that paper, careful consideration was given to issues such as the Tumor Microenvironment and the extracellular matrix (ECM). A two-dimensional regular grid was populated with healthy cells, proteins, an initial cancer cell, as well as fences bordering the ECM. The model forms the basis of this three-species model in which we can also consider the interactions with immune cells. We note that healthy cells, immune cells and the fixed number of ECM proteins are static, with only cancer cells moving. Every cancer cell and healthy cell has a stickiness value, and the jump radius of a cancer cell is the number of positions a cell can move at a time. Healthy cells (that have an associated age) have a probability of dividing when mature enough and can divide at most a fixed number of times. A key parameter is the fixed competition value between immune cells and cancer cells, so that when a cancer cell reaches a position that an immune cell occupies, both compete and only the more powerful cell remains, based on the generation of a random number that is compared with the competition value. When cancer cells divide, a daughter cell is placed as close to the mother cell as possible. When cancer cells reach the boundary of the domain, there is a probability of degrading the surrounding fence structure that can enable cancer cells to exit the domain, allowing for the possibility of metastasis. In Chapter 4 of Weerasinghe [9], Weerasinghe considers an initial exploration of the dynamics in which intervention by immune cells represents an immunotherapy treatment. Several authors have developed models, to study the effects of immunotherapy treatments; to those belong ODE-based models [10–12], as well as agent-based models [13–15].

In the agent-based model in Chapter 4, immune cells are placed in the outer region of the domain. The immune cells are immobile. Each immune cell has a competition rate. When a cancer cell or group of cancer cells (due to the stickiness of the cells) reaches a grid position occupied by an immune cell, they compete with one another, and the weaker cell dies. The competition value on $[0, 1]$ of a cancer cell $C_T$ is fixed, while the competitiveness rate of an immune cell, $C_I$, can be fixed or be uniformly drawn from the interval $[a, 1]$. The cancer cell will die if $C_T < C_I$. In Weerasinghe [9], values of $a = 0,\ 0.5$ and $0.75$ are considered. We note that no equation learning of an underlying ordinary differential equation is given in Weerasinghe [9], rather multiple simulations of the agent-based model are performed.

Three different strategies are considered for introducing immune cells into the agent-based model [9]: a one-off strategy, a repeated injection at fixed times, or injection when some percentage (such as 25%) of immune cells are lost. These are denoted as strategies 1, 2, and 3. Immune cells are initially injected when a certain number of cancer cells are detected (such as 500). Figure 2 (Figure 4.3 in Weerasinghe [9]) shows an initial progression of the cancer displaying healthy, cancer, immune cells, and ECM proteins (blue, red, purple, and black). Figure 4.5 in Weerasinghe [9] compares the dynamics for the three strategies when $a = 0$. Introducing immune cells keeps the healthy cell density at a higher value than without immune cells, and strategy two seems the most effective. Figure 4.14 in Weerasinghe [9] compares the dynamics with $a = 0, 0.5, 0.7$, and a higher value of $a$ seems to improve the effectiveness of strategies one and three. The immune cell model uses the same parameter values as in the two-species model (so that the initial healthy cell density is 0.324, the ECM protein density is 0.1, and the ECM breakdown probability is 0.5). The additional aspects involve the initial immune cell density and the nature of the competition between the immune cells and the cancer cells.

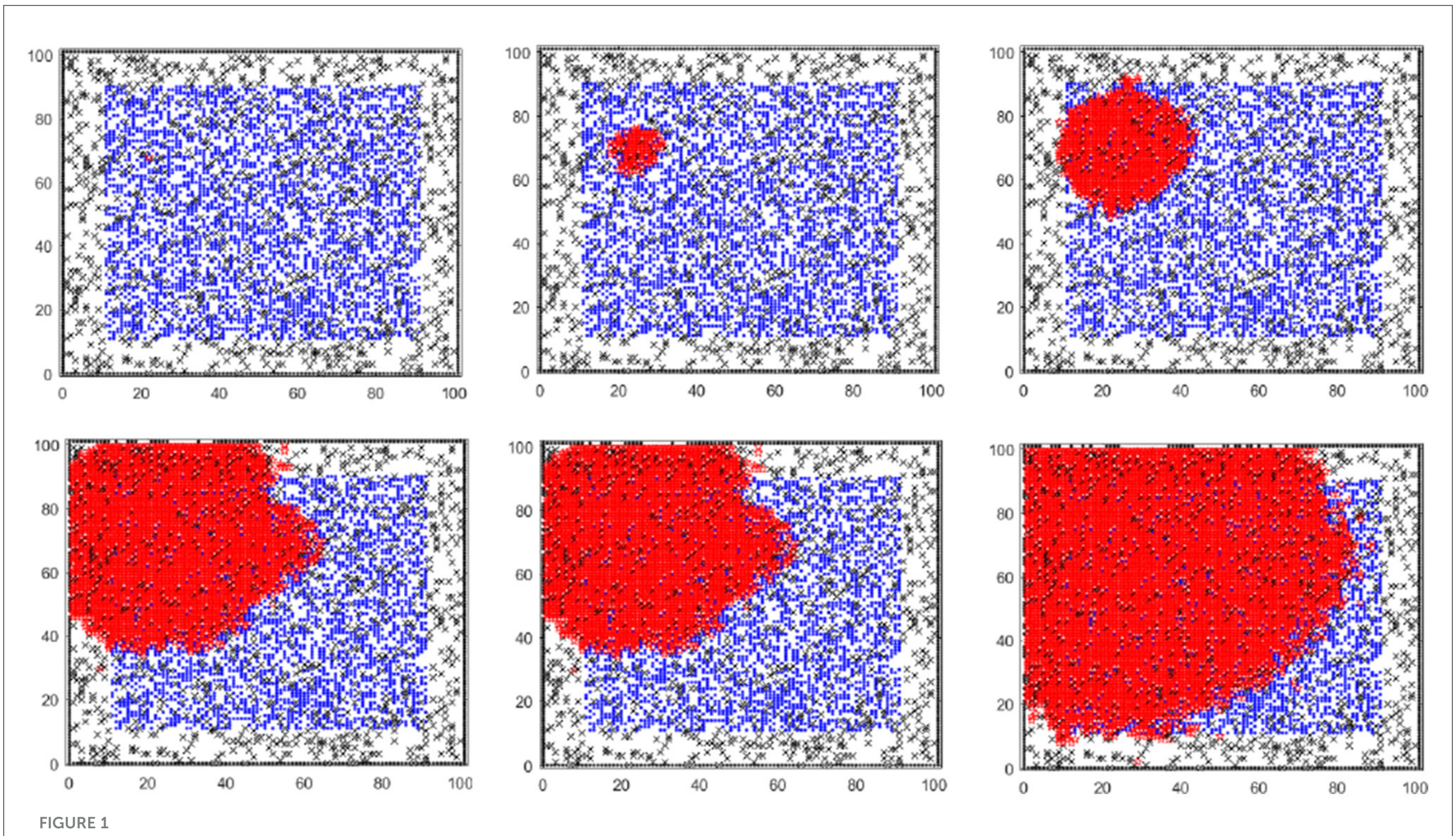

FIGURE 1
Cancer progression in the agent-based model: blue and red nodes represent healthy and cancer cells, respectively. Black nodes represent ECM proteins (obstacles) in the domain. Starting with one cancer cell, the cancer cell population increases, and the healthy cell population decreases with time.

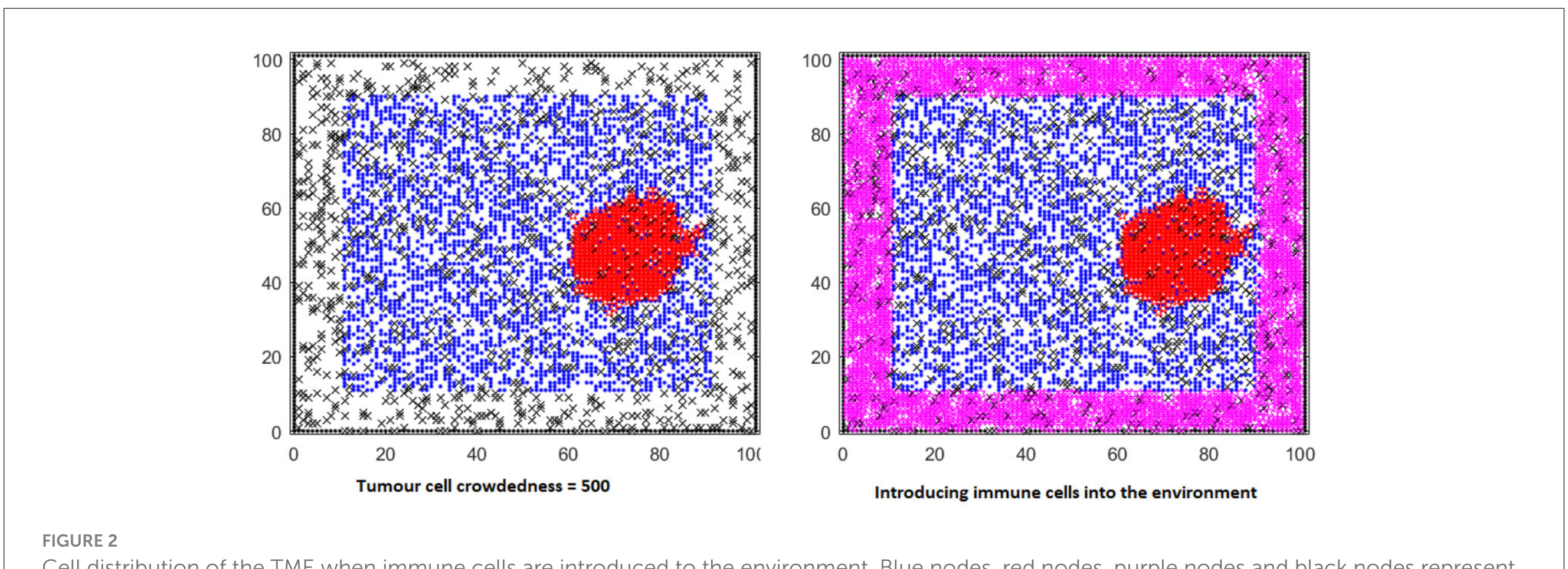

FIGURE 2
Cell distribution of the TME when immune cells are introduced to the environment. Blue nodes, red nodes, purple nodes and black nodes represent healthy cells, cancer cells, immune cells, and ECM proteins, respectively.

We will simulate the immune cell agent-based model with just a single injection of the immune cells and a fixed competition value between immune cells and cancer cells (cf. Section 2.4). The reason for adopting the single injection protocol is that this is clinically simpler to implement than other protocols and indeed, in the recently developed CAR-T immunotherapy protocol, a single injection of re-engineered immune cells is performed [16]. This will also simplify the use of equation learning to build a three-dimensional system of ordinary differential equations based on a library of interactions between the three species for a number of scenarios based on different initial conditions for the immune cells and a fixed competition value between immune cells and cancer cells.

## 2.2 Building a system of ODEs using Equation Learning

In Burrage et al. [6], we adapted the approach of Brunton and Kutz [5, 17] based on their sparse identification of nonlinear dynamics (SINDy) to discover an underlying ODE of our two species agent-based stochastic model [9] describing the interaction of healthy cells ($H$) and cancer cells ($C$). An issue with the standard SINDy approach is that the ODE that is constructed is done component by component, and so the ensuing system of ODEs has system components decoupled from one another. However, in Burrage et al. [6], the equation learning is based on a library of *chemical reactions* describing the evolution of the population counts of species $S_i$ (here, $H$ and $C$). Reactions take the form $l_i S_i \xrightarrow{k} r_i S_i$ with $k$ being the rate constant, $r_i, l_i \in \mathbb{N}$ the stoichiometric constants, and $\nu, \nu_i = r_i - l_i$ the stoichiometric coefficient vector. Under the Law of Mass Action, the rate constants are multiplied by the concentrations of the reactants. With this assumption, a library of functions can be constructed with the stoichiometric vectors coupling the components. The ODE that is built in this way can be interpreted as a set of reactions and respects the interactions inherent in the agent-based model. We built a library of chemical reactions based on $C$ and $H$ with 8 unimolecular reactions and 9 bimolecular reactions. The library of 17 possible reactions is

$$\theta(c,h) = \Bigg( \underset{k_1}{\begin{pmatrix} -c \\ 0 \end{pmatrix}}, \underset{k_2}{\begin{pmatrix} 0 \\ -h \end{pmatrix}}, \underset{k_3}{\begin{pmatrix} -c \\ c \end{pmatrix}}, \underset{k_4}{\begin{pmatrix} h \\ -h \end{pmatrix}}, \underset{k_5}{\begin{pmatrix} c \\ 0 \end{pmatrix}}, \underset{k_6}{\begin{pmatrix} 0 \\ c \end{pmatrix}}, \underset{k_7}{\begin{pmatrix} 0 \\ h \end{pmatrix}},$$

$$\underset{k_8}{\begin{pmatrix} h \\ 0 \end{pmatrix}}, \underset{k_9}{\begin{pmatrix} -c^2 \\ 0 \end{pmatrix}} \underset{k_{10}}{\begin{pmatrix} 0 \\ -h^2 \end{pmatrix}} \underset{k_{11}}{\begin{pmatrix} -\frac{1}{2}c^2 \\ 0 \end{pmatrix}} \underset{k_{12}}{\begin{pmatrix} -c^2 \\ \frac{1}{2}c^2 \end{pmatrix}} \underset{k_{13}}{\begin{pmatrix} 0 \\ -\frac{1}{2}h^2 \end{pmatrix}} \underset{k_{14}}{\begin{pmatrix} \frac{1}{2}h^2 \\ -h^2 \end{pmatrix}}$$

$$\underset{k_{15}}{\begin{pmatrix} -ch \\ -ch \end{pmatrix}} \underset{k_{16}}{\begin{pmatrix} 0 \\ -ch \end{pmatrix}} \underset{k_{17}}{\begin{pmatrix} -ch \\ 0 \end{pmatrix}} \Bigg)$$

with associated rate constants $k_1, \cdots, k_{17}$. For example, the vector number 3 corresponds to the reaction $C \xrightarrow{k_3} H$ and vector 12 to $2C \xrightarrow{k_{12}} H$.

The approach to find the non-negative rate constants is described extensively in Burrage et al. [6], but essentially we build data vectors $\tilde{C}$ and $\tilde{H}$ (computed as means over 500 simulations on a fixed time mesh). We then compute approximations to the derivatives $\tilde{C}'$ and $\tilde{H}'$ and solve a least squares problem

$$\min \left( \left\| \begin{pmatrix} \tilde{C}' \\ \tilde{H}' \end{pmatrix}^{\top} - \theta(\tilde{C}, \tilde{H})K \right\|_2^2 \right),$$

where $\theta(\tilde{C}, \tilde{H})$ is the library of chemical reactions evaluated at the data vectors and $K = (k_1, \cdots, k_{17})^{\top}$ is the vector of non-negative rate constants that is to be found. This can be solved by a non-negative least squares algorithm such as `lsqnonneg` in MATLAB. Once these rate constants have been found, we can use the Law of Mass Action to write out the ODE system. Now, the Law of Mass Action with $m$ reactions can be described by an ODE

$$y' = \sum_{j=1}^{m} \nu_j a_j(y), \quad (1)$$

with stoichiometric vectors $\nu_j$ and propensity functions $a_j(y)$. There is a well-established theory for discrete stochastic chemical kinetics—as the number of particles becomes large, we arrive at the ordinary differential equation regime described by the Law of Mass Action [18]. This regime arises in simple cases such as unimolecular and bimolecular reactions as well as for more complicated nonlinear reactions based around enzyme-like reactions. However, equation learning is much more difficult in this case and any analysis, such as equilibrium analysis, becomes much too complicated so we avoid this latter strategy.

In the case of the above library, the corresponding ODE after applying Equation 1 is

$$\begin{aligned} c' &= (k_5 - k_1 - k_3)c + (k_4 + k_8)h - (k_9 + \frac{1}{2}k_{11} + k_{12})c^2 \\ &\quad + \frac{1}{2}k_{14}h^2 - (k_{15} + k_{17})ch \\ h' &= (k_7 - k_2 - k_4)h + (k_3 + k_6)c - (k_{10} + \frac{1}{2}k_{13} + k_{14})h^2 \\ &\quad + \frac{1}{2}k_{12}c^2 - (k_{15} + k_{16})ch. \end{aligned}$$

We note from the library that there is a natural coupling between the two components from reactions 3, 4, 12, 14, and 15.

Within the agent-based model [6], the authors took a time step of 2/135 over a time interval of $26\frac{2}{3}$ that implies 1,800 steps. They also chose $\tilde{H}(0) = 0.324$ and $\tilde{C}(0) = 0.0001$. Better matches to the data seemed to be obtained when just sampling from 180 steps. Whether we used all 1,800 data points or sampled every 10 steps, the non-negative least squares algorithm determined that $k_1, k_2, k_3, k_4, k_6, k_8, k_{11}, k_{13}$ and $k_{16}$ were all zero, leaving only 8 reactions in the library [6].

It should be noted that this choice of reactions to be placed in the library can lead to issues of linear dependence between certain reactions—for example, $C + C \rightarrow 0$ and $C + C \rightarrow C$. In this case, `lsqnonneg` flags a highly ill-conditioned system but still converges to a solution in which $C + C \rightarrow 0$ appears in the final set, but $C + C \rightarrow C$ is removed.

## 2.3 Equation learning formulation for the augmented ABM

If we now consider the three-dimensional agent-based model with cancer cells, $C$, healthy cells, $H$, and immune cells, $I$, and let all the unimolecular and bimolecular reactions occur (with only two reactants on the left-hand side) then there are 27 unimolecular reactions and 33 bimolecular reactions—so a total of 60 reactions. Different from Burrage et al. [6], we avoid some collinearities without loss of generality by only considering reactions with a stoichiometric change of at most one per species, that is, $\nu_i \leq 1$ for $i \in \{C, H, I\}$. We can write these reactions as below.

Unimolecular (27):

$$
\begin{gathered}
C, H, I \to 0;\ C \to H, I;\ H \to C, I;\ I \to C, H \\
C \to C + C,\ C + H,\ C + I, H + I; \\
H \to H + C, H + H, H + I, C + I; \\
I \to I + C, I + I, I + H, C + H \\
0 \to C, H, I, C + H, C + I, H + I
\end{gathered}
$$

Bimolecular (33):

$$
\begin{gathered}
C + C \to C, C + H, C + I \\
C + H \to 0, C, H, I, C + C, C + I, H + H, H + I \\
C + I \to 0, C, H, I, C + C, C + H, I + H, I + I \\
H + H \to H, H + C, H + I \\
H + I \to 0, C, H, I, H + C, H + I, H + H, I + I \\
I + I \to I, I + C, I + H
\end{gathered}
$$

Further, from the definition of the ABM we can infer that immune cells only interact with cancer cells. Incorporating this prior knowledge, we may work with a reduced library of 49 reactions, where reactions of the form $I + I \to *$ and $H + I \to *$ are removed (with the $*$ signifying "any" combination of products).

For the three-species ABM, the least squares problem is

$$
\min \left( \left\| \begin{pmatrix} \tilde{C}' \\ \tilde{H}' \\ \tilde{I}' \end{pmatrix} - \theta(\tilde{C}, \tilde{H}, \tilde{I})\, K \right\|_2^2 \right), \tag{2}
$$

where $K$ is a vector of length 60 (49).

We let the approach learn which of the 60 (49) reactions appear in terms of two key parameters—namely $\tilde{I}(0)$ and the fixed competition parameter between cancer cells and immune cells. Note that $\tilde{I}(0)$ is considered as a parameter of the model rather than an initial condition, as it is varied as part of the treatment strategies and may even be time-dependent.

## 2.4 Learning equations for the augmented agent-based model

Solving Equation 2 requires two main ingredients: time-series data $\tilde{C}, \tilde{H}, \tilde{I}$ on the amount of $C$, $H$, and $I$, as well as the derivatives $\tilde{C}'$, $\tilde{H}'$, and $\tilde{I}'$. In our case, the former is obtained by simulating the three-species ABM for 20 time units and taking the mean concentrations over 100 stochastic replications (cf. Figure 3). The measurement step is set to 0.01 time units, resulting in (mean) time-series data for C, H, and I with 2,000 measurements each. The derivatives can be determined by numerical differentiation with central differences. However, the solution to Equation 2 is very sensitive to the accuracy of these derivatives, as they form the target vector. We also compare this to a numerical differentiation scheme employing Kalman-filtered derivatives.[1]

Further, as was demonstrated in Burrage et al. [6], the number of measurements in $\tilde{C}$, $\tilde{H}$, and $\tilde{I}$ can considerably influence the solution of Equation 2. Thus, we also vary the sampling interval, using only every 10th, 20th, 40th, or 50th data point.

Finally, there are typically a lot of different solutions to the inverse problem of determining a model (structure and parameters) from data. Thus, without additional information, it is generally impossible to determine a single "correct" result. For example, as also found in Burrage et al. [6] and discussed in Section 2.2, collinearities pose a problem: The reactions $\xrightarrow{2k} C$ and $\xrightarrow{k} 2C$ have the same effect (considering ODE semantics), so choosing among them is impossible without additional knowledge. This also impacts the ability of linear solvers to find an adequate solution, as the equation system (Equation 2) becomes ill-conditioned. Thus, it is helpful to incorporate prior knowledge. To avoid collinearities of reactions as shown above, then for the three species ABM we only consider reactions with stoichiometry one (requiring $\nu_i \leq 1$ for the change of each species $i$), leading to 60 reactions (cf. Section 2.3). We call this the *complete library* for our problem formulation. Additionally, we can exclude interactions of immune cells with healthy cells and among immune cells themselves, which leads to a *constrained library* of 49 reactions. The necessity of incorporating constraints and prior knowledge is also highlighted in Santosa and Weitz [8] where the authors demonstrate identifiability issues in equation learning (even before SINDy was introduced) associated with sparsity-enforcing approaches when minimizing $\ell_1$ or $\ell_0$ norms.

### 2.4.1 A single reaction model reflecting ABM parameters

We solve Equation 2 using the `nnls` optimizer from the Python package `scipy` [19] for the data obtained from the different combinations for the initial percentage of immune cells ($\tilde{I}(0) \in \{0.05, 0.2, 0.1\}$) and the competition value between cancer and immune cells ($C_I \in \{0.5, 0.75\}$). The result of the equation learning is then a set of six models, one per parameter combination in $\mathrm{dom}(\tilde{I}(0)) \times \mathrm{dom}(C_I)$. We hypothesize that these six models share a large part of their reactions, as they are based on data from the same ABM. A final model is then determined by taking the union over the sets of reactions included in all six scenarios. If our hypothesis holds, this union model should contain much less than the 60 (complete library) or 49 (constrained library) reactions in the library. Taking the union instead of the intersection acknowledges the fact that, in some scenarios, the rate constant of a specific reaction might be (very close to) zero, while it is important in many others. Further, there should be a relation between the two varied parameters of the ABM and the rate constants and

1 see https://github.com/hugohadfield/kalmangrad.

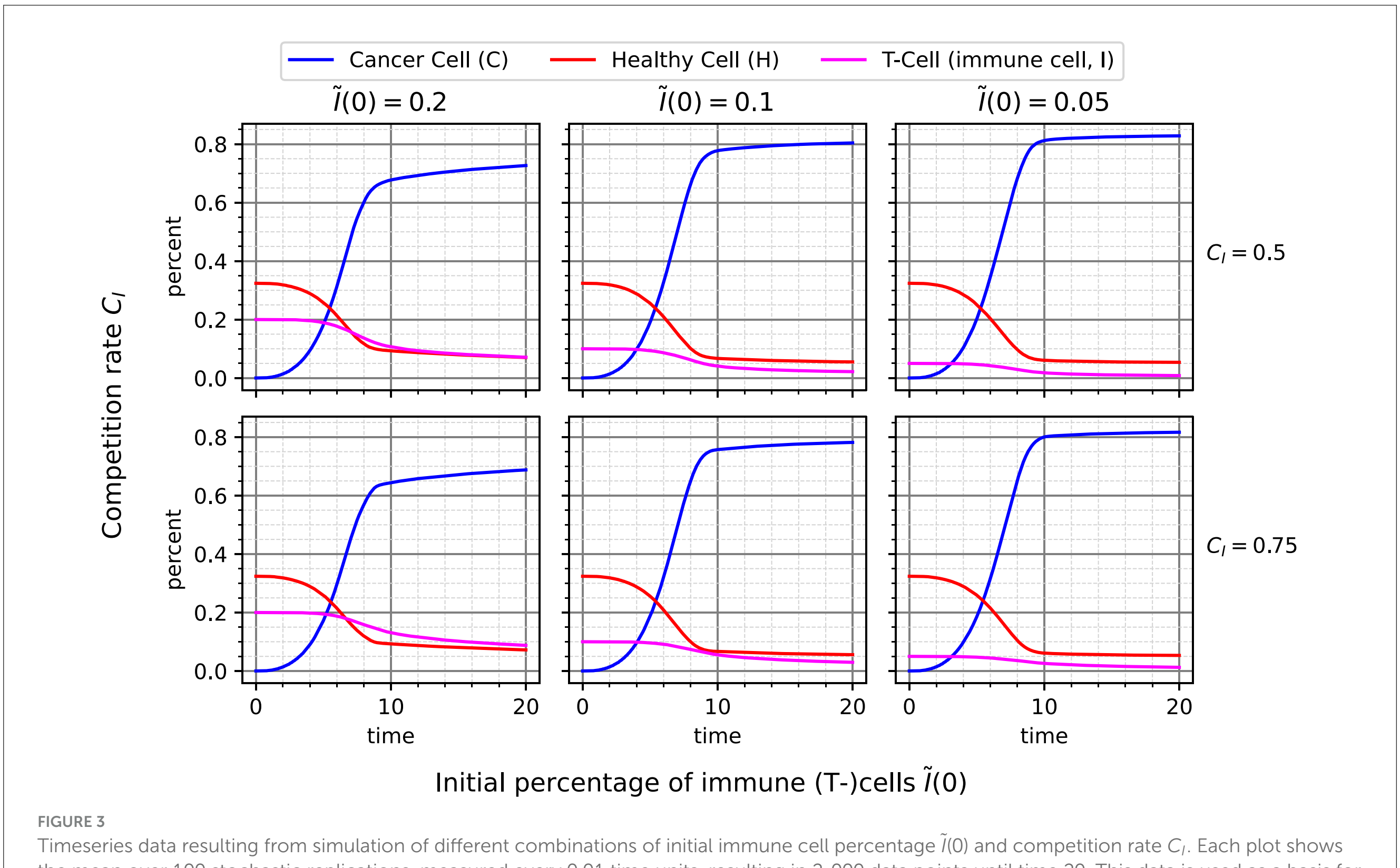

FIGURE 3
Timeseries data resulting from simulation of different combinations of initial immune cell percentage $\tilde{I}(0)$ and competition rate $C_I$. Each plot shows the mean over 100 stochastic replications, measured every 0.01 time units, resulting in 2,000 data points until time 20. This data is used as a basis for learning the reaction model.

reactions included in the union model. Exploiting this relation should enable approximating a wide variety of behaviors of the ABM and drawing conclusions on the ABM's properties from analyzing the union model.

A union model's fidelity to the ABM can be tested a priori, without knowledge of this relation, by trying to fit it to the six scenarios again. For this, we again use Equation 2, but limit the library to the reactions contained in the union model. We integrate the calibrated models using LSODA and regard the mean over the root mean squared errors (RMSEs) with respect to the six time series as the union model's accuracy. This measure allows us to compare different values for the hyperparameters involved in equation learning.

### 2.4.2 Hyperparameter scan

As we outlined above, learning reaction models involves some choices that can strongly influence the resulting model. These include the selection of reactions in the library (complete/constrained), the method used for numerical differentiation of the time series data (central differences on filtered/unfiltered data), and the measurement interval (considering only every n-th point from the data with $n \in \{10, 20, 40, 50\}$). We determine a union model as outlined above for every one of the above 16 combinations of hyperparameters, selecting the model with the smallest RMSE as the final result. The complete results of this analysis are visualized in Figures 4, 5. The former shows the coefficient values obtained for each reaction in our complete library when varying the sampling frequency $n$ of the data (y-axis) and the filtering for differentiation (x-axis). Each subplot depicts the library of reactions on the x-axis and the ABM parameter configurations on the y-axis. Figure 5 shows the same for the case of the constrained library. Some interesting observations can be made regarding the choice of hyperparameters. When using filtered gradients (right columns in the figures), the magnitude of some coefficients decreases with increasing value of $n$, such as for the reaction $2H \rightarrow H$ in Figure 4. On the other hand, when using unfiltered gradients, this effect cannot be observed. Looking at the behavior over ABM parameter configurations (vertical streaks), we can observe that, indeed, there is a relation between ABM parameter configurations to the coefficients of many reactions. For example, in Figure 4, bottom left, the coefficient of the reaction $2I \rightarrow I$ is always higher for $C_I = 0.5$ than for $C_I = 0.75$. Further, it seems to increase when $I(0)$ increases (this can be seen by considering every second cell from bottom to top). Such patterns can be observed for many of the other reactions in both, the complete and constrained, cases.

## 3 Results

### 3.1 Equations for the three-species agent-based model

Table 1 summarizes the results of the hyperparameter scan. With the constrained library, where immune cells are not allowed

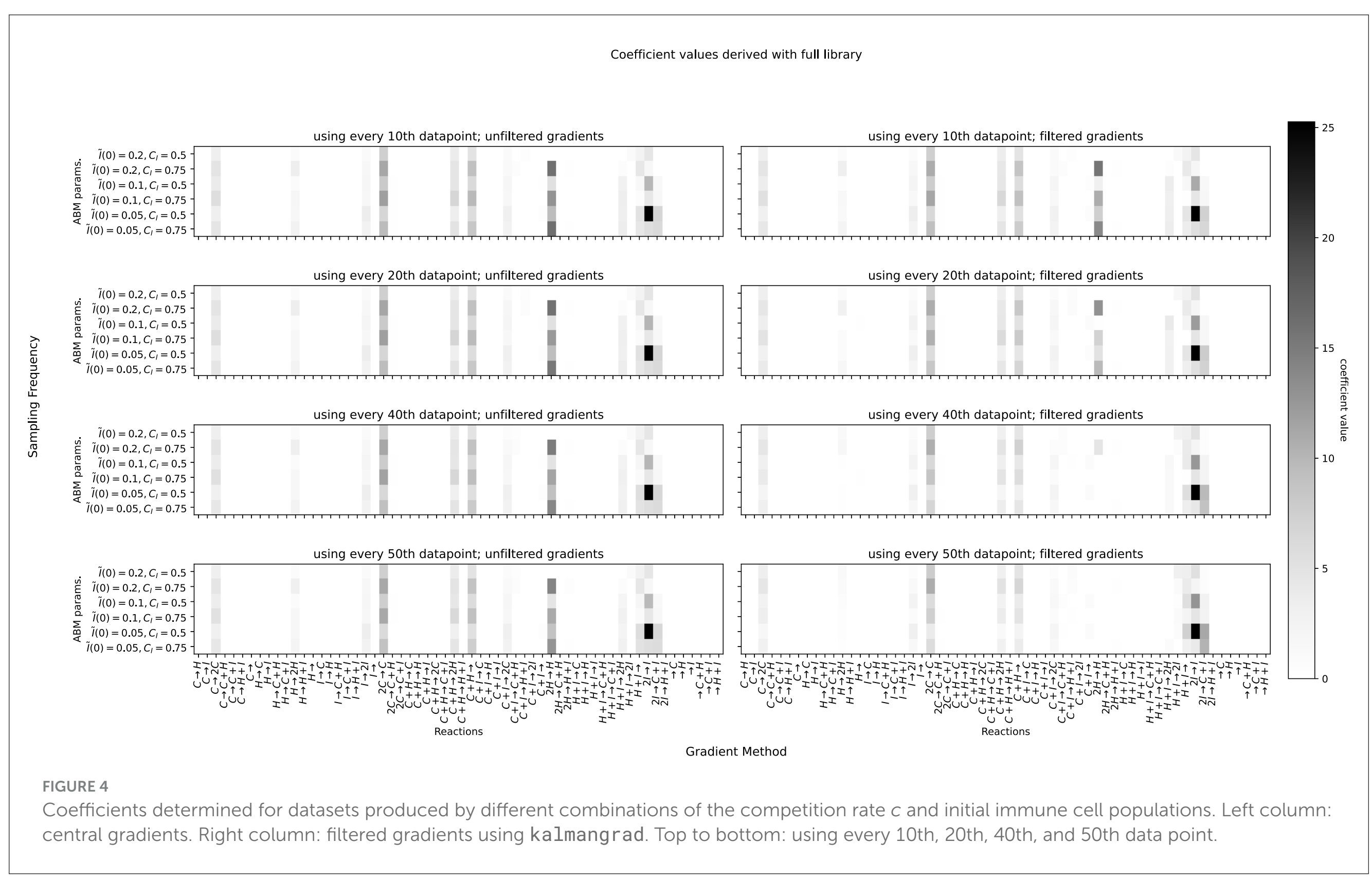

FIGURE 4
Coefficients determined for datasets produced by different combinations of the competition rate $c$ and initial immune cell populations. Left column: central gradients. Right column: filtered gradients using `kalmangrad`. Top to bottom: using every 10th, 20th, 40th, and 50th data point.

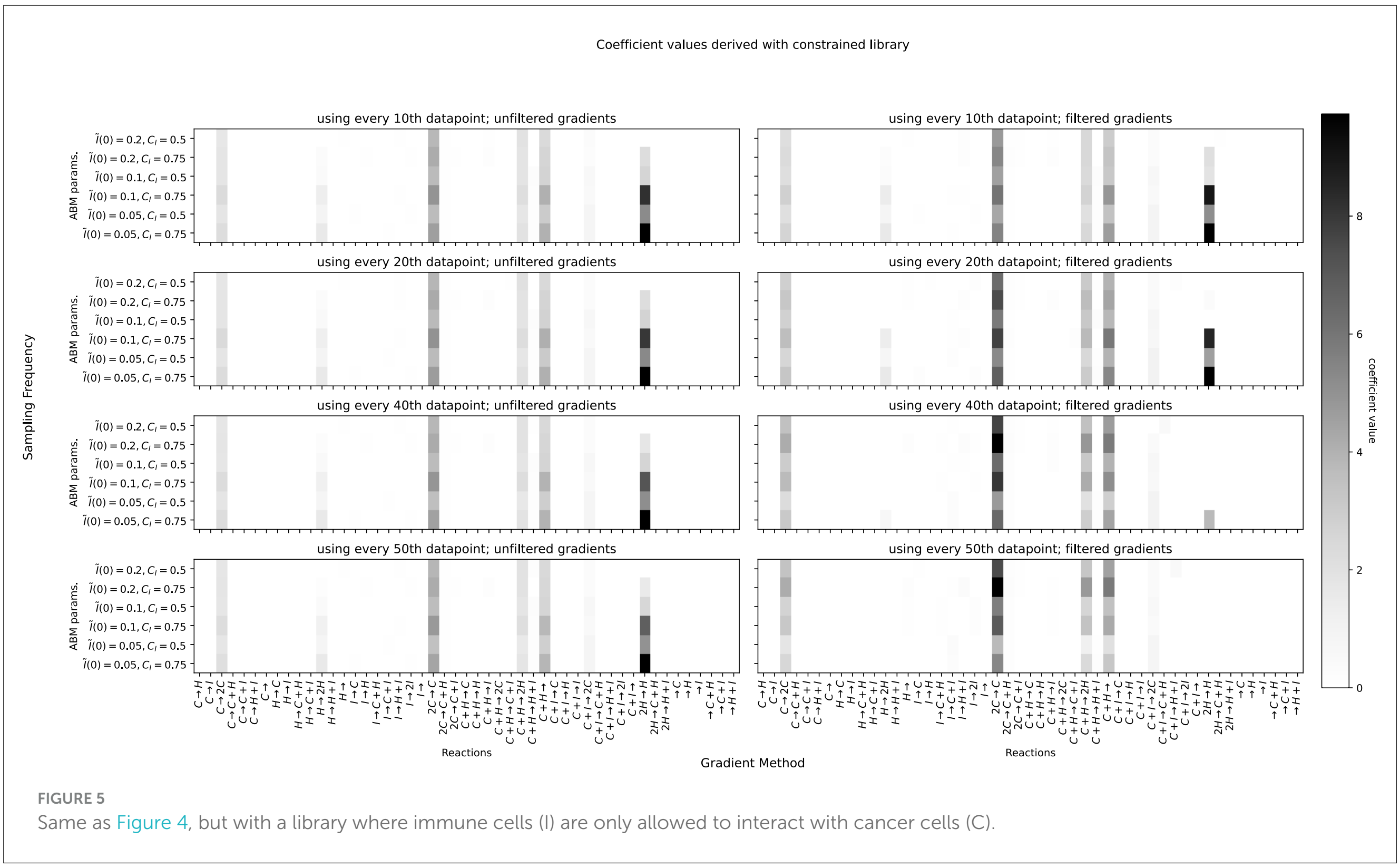

FIGURE 5
Same as Figure 4, but with a library where immune cells (I) are only allowed to interact with cancer cells (C).

to interact with healthy cells or themselves, the smallest mean over RMSEs ($\approx 5.45 \times 10^{-3}$) was obtained when using unfiltered central difference gradients over every 10th data point, leading to 12 reactions. In the unconstrained case, a slightly lower mean over RMSEs ($\approx 5.06 \times 10^{-3}$) could be obtained using the same hyperparameters, leading to 14 reactions. We choose

TABLE 1 Results of the hyperparameter scan.

| Constrained? | $n$ | Gradient method | No. reactions | Mean RMSE (data) | Mean RMSE (der.) |
|---|---|---|---|---|---|
| **True** | **10** | **Unfiltered** | **12** | $\mathbf{5.45 \times 10^{-3}}$ | $\mathbf{108 \times 10^{-3}}$ |
| True | 10 | Filtered | 12 | $6.46 \times 10^{-3}$ | $94.8 \times 10^{-3}$ |
| True | 20 | Unfiltered | 12 | $5.72 \times 10^{-3}$ | $76.4 \times 10^{-3}$ |
| True | 20 | Filtered | 12 | $7.69 \times 10^{-3}$ | $59.6 \times 10^{-3}$ |
| True | 40 | Unfiltered | 12 | $6.00 \times 10^{-3}$ | $53.1 \times 10^{-3}$ |
| True | 40 | Filtered | 12 | $9.93 \times 10^{-3}$ | $37.6 \times 10^{-3}$ |
| True | 50 | Unfiltered | 12 | $6.01 \times 10^{-3}$ | $46.7 \times 10^{-3}$ |
| True | 50 | Filtered | 11 | $10.8 \times 10^{-3}$ | $34.7 \times 10^{-3}$ |
| **False** | **10** | **Unfiltered** | **14** | $\mathbf{5.06 \times 10^{-3}}$ | $\mathbf{84.8 \times 10^{-3}}$ |
| False | 10 | Filtered | 14 | $6.11 \times 10^{-3}$ | $76.6 \times 10^{-3}$ |
| False | 20 | Unfiltered | 14 | $5.31 \times 10^{-3}$ | $60.0 \times 10^{-3}$ |
| False | 20 | Filtered | 14 | $7.41 \times 10^{-3}$ | $50.5 \times 10^{-3}$ |
| False | 40 | Unfiltered | 14 | $5.59 \times 10^{-3}$ | $41.7 \times 10^{-3}$ |
| False | 40 | Filtered | 13 | $9.64 \times 10^{-3}$ | $35.3 \times 10^{-3}$ |
| False | 50 | Unfiltered | 14 | $5.63 \times 10^{-3}$ | $36.5 \times 10^{-3}$ |
| False | 50 | Filtered | 13 | $10.7 \times 10^{-3}$ | $33.5 \times 10^{-3}$ |

Hyperparameter values, resulting number of reactions, mean over RMSEs (with respect to fitting the union model to the ABM trajectories), and mean over RMSEs (with respect to numerically determined gradients). Bold lines indicate the hyperparameter combinations leading to the most accurate models with respect to the data when using the unconstrained and constrained library, respectively (cf. Section 3.1).

to continue with the model based on the constrained library, for being more parsimonious and also recognizing our prior knowledge about possible reactions while achieving a very similar accuracy as the unconstrained model. In Figure 5, this combination is shown on the top left. The reactions of our chosen union model are:

$$\begin{array}{rlrl} C \xrightarrow{k_0} & C+C, & H \xrightarrow{k_1} & H+H, \\ I \xrightarrow{k_3} & C+I, & C+C \xrightarrow{k_4} & C, \\ C+C \xrightarrow{k_6} & C+I, & C+H \xrightarrow{k_7} & H+H, \\ C+H \xrightarrow{k_9} & 0, & C+I \xrightarrow{k_{10}} & C+C, \\ I \xrightarrow{k_2} & C, & C+C \xrightarrow{k_5} & C+H, \\ C+H \xrightarrow{k_8} & H+I, & H+H \xrightarrow{k_{11}} & H. \end{array} \tag{3}$$

With the Law of Mass Action, we obtain the following differential equation system

$$\begin{aligned} c' &= k_0 c + (k_2 + k_3) i - \frac{1}{2}(k_4 + k_5 + k_6) c^2 \\ &\quad + (-k_7 - k_9) ch + k_{10} ci \\ h' &= k_1 h + \frac{1}{2} k_5 c^2 + (k_7 - k_9) ch - \frac{1}{2} k_{11} h^2 \\ i' &= -k_2 i + \frac{1}{2} k_6 c^2 + k_8 ch - k_{10} ci. \end{aligned} \tag{4}$$

A coupling of the derivatives can be observed over the coefficients $k_2, k_5, k_6, k_7, k_9$, and $k_{10}$.

Figure 6 shows the difference between the original data and fitting this best model (adjusting $k_0, \ldots, k_{11}$) to the six scenarios. It can be seen that the 12 learned reactions above can very accurately be fitted to the different ABM parametrizations and reproduce the ABM's results. We can now use the coupled equations from Equation 4 as a surrogate to analyse properties of the ABM, such as the equilibrium states.

## 3.2 Equilibrium analysis of the equation-learning ODEs

In this subsection, we study the equilibrium of the 6 ODEs constructed from the scenario data. We note from Equation 4 that each ODE takes the form (with the 13 coefficients $\theta_1, \ldots, \theta_5; \phi_1, \ldots, \phi_4; \lambda_1, \ldots, \lambda_4$ determined)

$$C' = \theta_1 C + \theta_2 I + \theta_3 C^2 + \theta_4 CH + \theta_5 CI \tag{5}$$

$$H' = \phi_1 H + \phi_2 C^2 + \phi_3 CH + \phi_4 H^2 \tag{6}$$

$$I' = \lambda_1 I + \lambda_2 C^2 + \lambda_3 CH + \lambda_4 CI. \tag{7}$$

We solve the right-hand side of each equation to 0. Studying each in turn, Equation 7 can be written as

$$I\,(\lambda_1 + \lambda_4 C) = -C(\lambda_3 H + \lambda_2 C) \tag{8}$$

or

$$I = -Cp/q, \quad p = \lambda_3 H + \lambda_2 C, \quad q = \lambda_1 + \lambda_4 C. \tag{9}$$

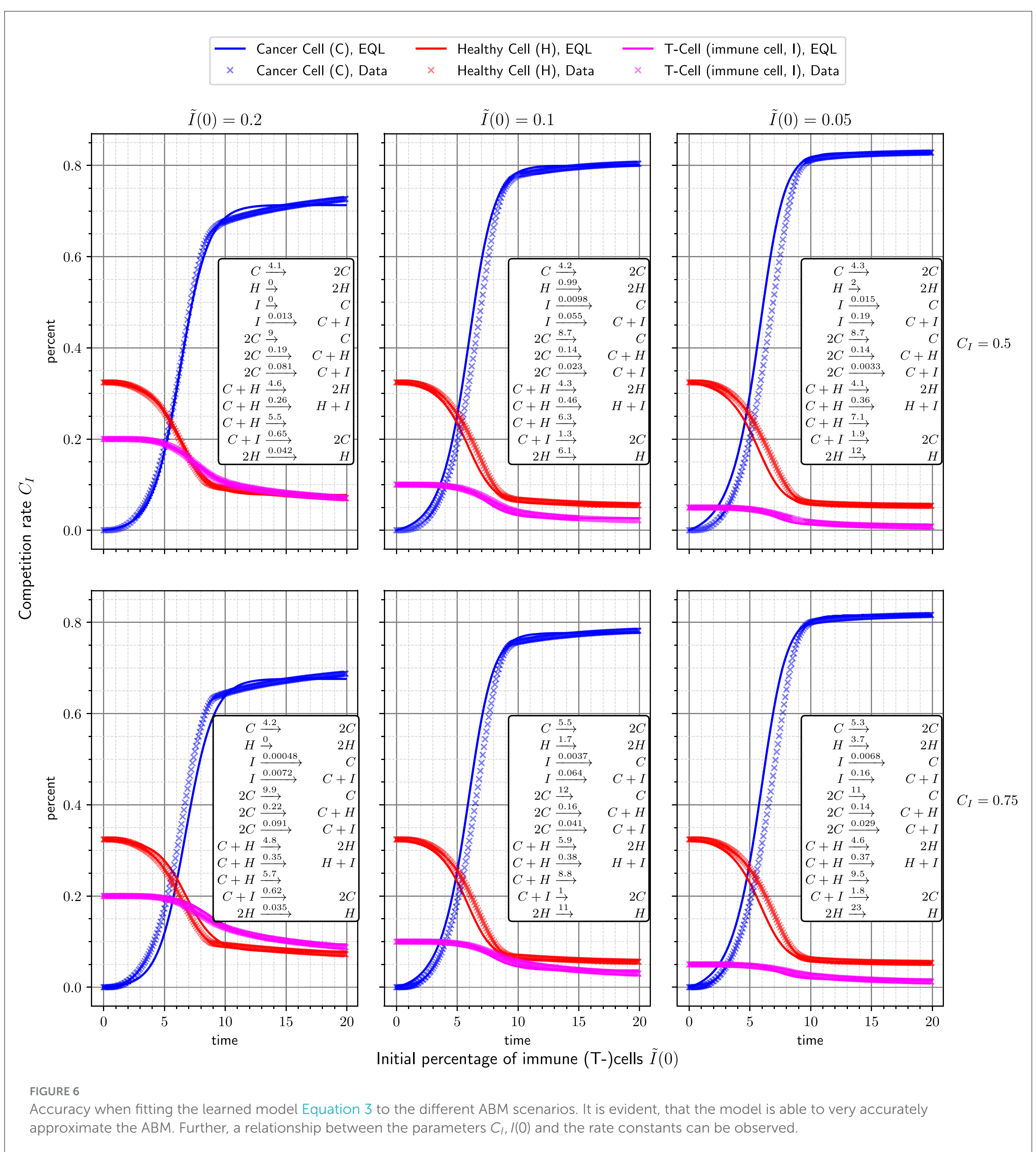

FIGURE 6
Accuracy when fitting the learned model Equation 3 to the different ABM scenarios. It is evident, that the model is able to very accurately approximate the ABM. Further, a relationship between the parameters $C_I$, $I(0)$ and the rate constants can be observed.

Now Equations 5, 9 lead to

$$C\,(\theta_1 + \theta_3 C + \theta_4 H) = C\frac{p}{q}(\theta_2 + \theta_5 C)$$

and with $C \neq 0$ yields

$$q\,(\theta_1 + \theta_3 C + \theta_4 H) = p\,(\theta_2 + \theta_5 C). \tag{10}$$

Finally Equation 6 is just

$$\phi_1 H + \phi_2 C^2 + \phi_3 CH + \phi_4 H^2 = 0 \tag{11}$$

so Equations 9–11 give

$$(\lambda_1 + \lambda_4 C)(\theta_1 + \theta_3 C + \theta_4 H) = (\lambda_3 H + \lambda_2 C)\,(\theta_2 + \theta_5 C)$$

or

$$(\lambda_1 + \lambda_4 C)(\theta_1 + \theta_3 C) - \lambda_2 C(\theta_2 + \theta_5 C) = H\,\big(\lambda_3(\theta_2 + \theta_5 C) \\ -\theta_4(\lambda_1 + \lambda_4 C)\big)\,. \tag{12}$$

TABLE 2 Cell densities at equilibrium.

| ODE | $C_I$ | $\tilde{I}(0)$ | $\tilde{C}$ | $\tilde{H}$ | $\tilde{I}$ |
|---|---|---|---|---|---|
| 1 | 0.5 | 0.2 | 0.7313 | 0.0787 | 0.0771 |
| 2 | 0.5 | 0.1 | 0.8117 | 0.0571 | 0.0293 |
| 3 | 0.5 | 0.05 | 0.8364 | 0.0536 | 0.0108 |
| 4 | 0.75 | 0.2 | 0.6763 | 0.0798 | 0.0946 |
| 5 | 0.75 | 0.1 | 0.7838 | 0.0558 | 0.0371 |
| 6 | 0.75 | 0.05 | 0.8319 | 0.0506 | 0.0170 |

Write Equation 12 as

$$H = \frac{R}{S} \tag{13}$$

$$\left.\begin{aligned} R &= \lambda_1\theta_1 + C(\lambda_1\theta_3 + \lambda_4\theta_1 - \lambda_2\theta_2) + C^2(\lambda_4\theta_3 - \lambda_2\theta_5) \\ S &= \lambda_3\theta_2 - \lambda_1\theta_4 + C(\lambda_3\theta_5 - \lambda_4\theta_4). \end{aligned}\right\} \tag{14}$$

Substitute Equation 13 into Equation 11, then

$$\phi_1\frac{R}{S} + \phi_2 C^2 + \phi_3 C\frac{R}{S} + \phi_4\frac{R^2}{S^2} = 0$$

or

$$\phi_1 RS + \phi_2 C^2 S^2 + \phi_3 CRS + \phi_4 R^2 = 0. \tag{15}$$

Writing (and comparing coefficients with Equation 14)

$$\begin{aligned} R &= \alpha_1 + \alpha_2 C + \alpha_3 C^2 \\ S &= \beta_1 + \beta_2 C \end{aligned}$$

and substituting into Equation 15 gives a fourth degree polynomial in $C$:

$$E_4C^4 + E_3C^3 + E_2C^2 + E_1C + E_0 = 0. \tag{16}$$

We will solve Equation 16 for $C$, Equation 13 for $H$, Equation 8 for $I$. It can be shown that

$$\begin{aligned} E_4 &= \phi_4\alpha_3^2 + \phi_3\alpha_3\beta_2 + \phi_2\beta_2^2 \\ E_3 &= \phi_1\alpha_3\beta_2 + 2\phi_2\beta_1\beta_2 + \phi_3(\alpha_3\beta_1 + \alpha_2\beta_2) + 2\phi_4\alpha_2\alpha_3 \\ E_2 &= \phi_1(\alpha_3\beta_1 + \alpha_2\beta_2) + \phi_2\beta_1^2 + \phi_3(\alpha_1\beta_2 + \alpha_2\beta_1) \\ &\quad + \phi_4(2\alpha_1\alpha_3 + \alpha_2^2) \\ E_1 &= \phi_1(\alpha_1\beta_2 + \alpha_2\beta_1) + \phi_3\alpha_1\beta_1 + 2\phi_4\alpha_1\alpha_2 \\ E_0 &= \phi_1\alpha_1\beta_1 + \phi_4\alpha_1^2. \end{aligned}$$

Some analysis (not given here) shows that there is only one positive zero for $C$ given by Equation 16 for each of the six scenarios. The results are given in Table 2.

The use of equilibrium analysis is two-fold. First, it allows us to see the long-term stability of cancer cells within the spatial domain, and secondly, it allows us to infer a specific relationship between the equilibrium value of cancer cells and the immune cell injection. This is vital in giving clinicians a clue as to what sort of initial immune cell injection value should be chosen in order to control the cancer proliferation to an appropriate value.

# 4 Discussion

## 4.1 Insights from the agent-based model and the learned equations

We adapted the two-species agent-based model describing the interaction between cancer and healthy cells developed in Weerasinghe et al. [1] to a three-species model that included immune cells. The aim of this was to probe the role of immunotherapy in cancers, such as multiple myeloma, based on a simple but effective stochastic spatial model. This was first developed in Weerasinghe [9]. Our model is different from that in that there is a single injection of immune cells into the spatial model, and there is a constant competition parameter between cancer cells and immune cells.

We use equation learning to construct a three-dimensional ODE model, thus removing space and stochasticity from the study. The main focus is then to probe the role of immune cells in dealing with the proliferation of cancer cells.

We have only simulated six scenarios and studied the six learned ODEs, as described in the previous section, in terms of a key characteristic, namely their steady-state behavior. Figure 7 shows that for the two competition values of 0.5 and 0.75, we get more or less a linear relationship between the steady-state concentration of the cancer cells and the initial concentration of the immune cells $\tilde{I}(0)$. Thus, for a given value of $C_I$ we would expect as we increase $\tilde{I}(0)$ that we would still maintain this linear relationship. This suggests that we do not need to run more expensive simulations of the agent-based model in order to learn additional features of the steady-state behaviors. Furthermore, comparisons of the concentrations of $C$, $H$, and $I$ at $T = 20$ for each of the six learned ODEs and the steady state values suggest that we do not need to run the agent-based model over very long time intervals to find steady-state values.

Figure 7 suggests that for a small value of the initial immune concentration $\tilde{I}(0)$, there is little effect of the competition value on the cancer cell concentration—recall that a higher value of $C_I$ (the competition value between cancer and immune cells), the more effective are the immune cells in suppressing the concentration of the cancer cells. Figure 7 also suggests that as we increase $\tilde{I}(0)$, the competition value plays an important role in suppressing the cancer cells. Furthermore, Figure 7 also shows that if we want to reduce the cancer cell concentration at steady-state, then engineering the immune cells so that they have a higher competition value and increasing the initial concentration of immune cells both play important roles.

In order to explore further the linear relationship between $C$ and $\tilde{I}(0)$ for both competition values $C_I = 0.5,\ 0.75$ we find the linear least squares solution in both cases. These are

$$\begin{aligned} C &= 0.8766 - 0.7154\,\tilde{I}(0) \\ C &= 0.8857 - 1.0427\,\tilde{I}(0). \end{aligned}$$

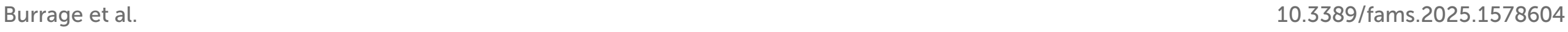

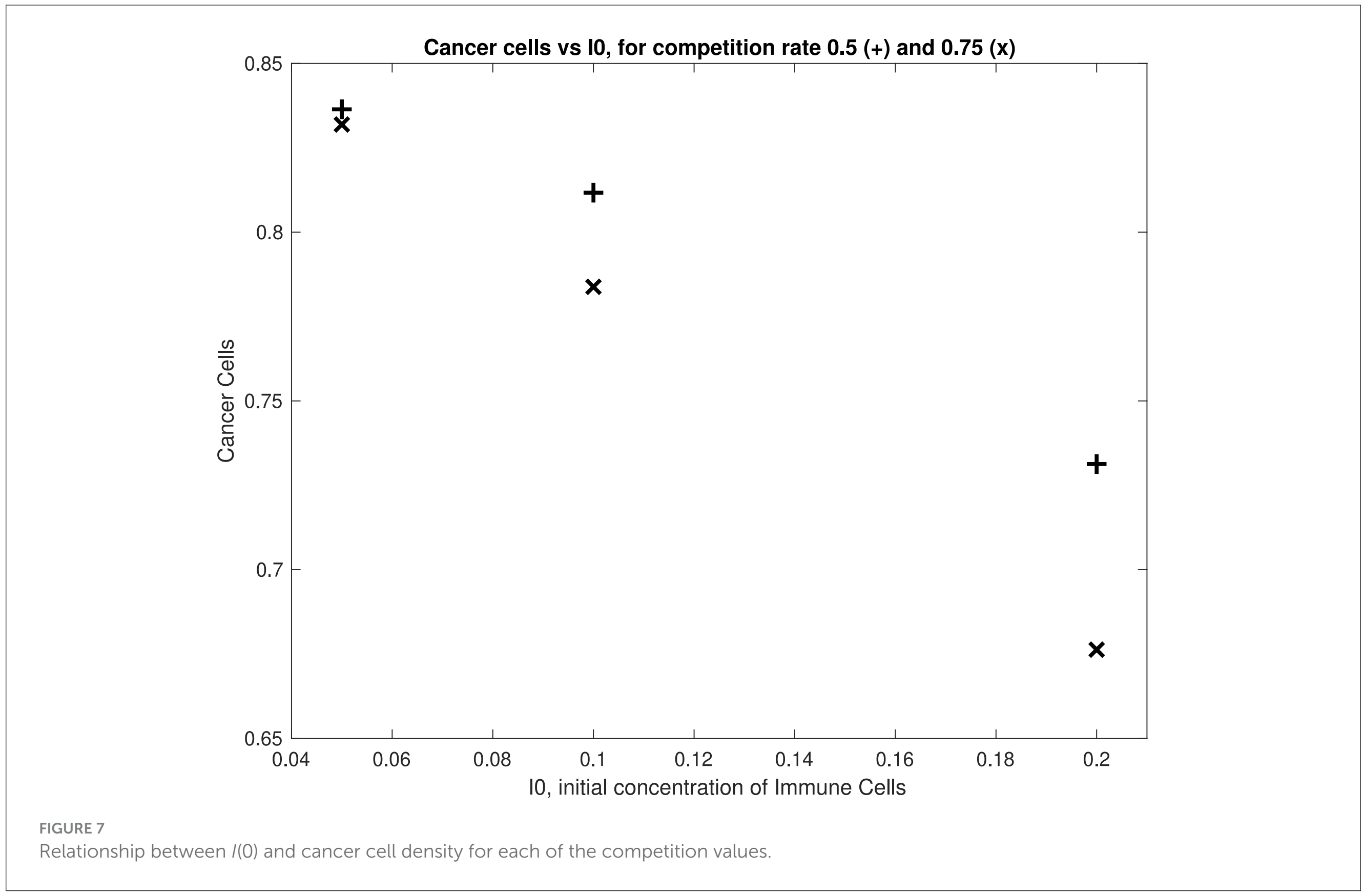

FIGURE 7
Relationship between $I(0)$ and cancer cell density for each of the competition values.

We can then ask, for example, what value of $\tilde{I}(0)$ gives a value of $C = 0.2$ at equilibrium. It turns out to be

$$\tilde{I}(0) = 0.9457, \quad C_I = 0.5$$
$$\tilde{I}(0) = 0.6576, \quad C_I = 0.75.$$

Of course, we can use this approach to study immunotherapy scenarios based on reducing the cancer to manageable levels over certain periods of time.

## 4.2 Related work

ABMs have become an established approach in cancer research [20, 21]. Modeling individuals and their interactions flexibly supports incorporating detailed biological mechanisms (also spatially and temporally constrained) at the micro level and studying emergent properties at the macro level. Many multiscale models exist that combine cellular behavior with intra-cellular behavior, such as signaling pathways [20, 21].

In a recent review of mechanistic learning in oncology [22], which contrasts knowledge-based and data-driven modeling, various approaches of their combination are listed, including equation learning. It is noted that "despite remarkable success in physics, symbolic regression applications in oncology are still scarce." One approach is Brummer et al. [23]. It uses a combination of sparse regression (SINDy) and dynamic mode decomposition (DMC) to estimate a system of ODEs from *in vitro* CAR T-cell glioma therapy data. The starting point is *in-vitro* data with high temporal resolution instead of an agent-based simulation model. Also, instead of deriving one model (in our case, a union of learned models) to capture all observed dynamics, in Brummer et al. [23], three different models are derived based on the different experimental settings of the *in-vitro* experiments. Our work thus adds to the currently very small body of literature on applying equation learning in oncology. We have shown that equation learning may not only be used to analyse data but also to simplify the analysis of existing (agent-based) models.

When looking more broadly at the application field of cell biology, the situation is slightly different. Discovering ODEs for biochemical systems from data is discussed in Daniels and Nemenman [24]. The authors automatically discover parsimonious models by systematically fitting models of increasing complexity. Several approaches [6, 25] are based on applying the seminal Sparse Identification of Nonlinear Dynamics (SINDy) [5] to derive ODEs.

Other approaches are based on the specification of biochemical systems as a set of reactions. One way to achieve this is by learning ODEs constrained by a specific coupling [6, 7, 26], as used here. Other approaches include stochastic heuristic search [27] and the chemical reaction neural network (CRNN) [28]. All methods mentioned above only consider the deterministic semantics of reaction models. When stochastic data is available, it can also be used to infer CRNs that allow for a stochastic interpretation, such as [29, 30].

The goal of these approaches is to learn entire simulation models based on data. Ideally, the data stem from *in-vitro* or *in-vivo* experiments. However, simulation can be used to generate synthetic data for machine learning in general (thus closing possible gaps in existing data) [31], and for evaluating equation learning in particular, as it allows us to directly compare the learned model with the ground truth with the simulation model generating the data [26].

We show that the approach [6] can generate an ODE model that reproduces the data of the more complex agent-based tumor model. As a next step, the equation model helps us derive properties of the simulation model and reduce the computational costs in experimenting with this original simulation model. The learned equations can be viewed as a surrogate [32] for the real model. However, unlike surrogate models, which are typically black box models such as artificial neural networks [32], the derived equation models can be interpreted by humans. In addition, the entire tooling for analyzing ODEs is available to assess diverse properties and behaviors of the equations approximating the agent-based simulation model, which we showed exemplarily based on a steady-state analysis.

## 5 Conclusions

We have seen that our approach of learning an ODE system from just a few agent-based modeling scenarios (here, six) and then using the ODE to study a key characteristic of the long-term cancer dynamics is very powerful. In particular, by establishing a linear relationship between cancer concentrations and the initial concentration of the immune cells, this enables us to make a simple analysis on the values of the competition and the initial immune concentration in order to reduce the cancer concentrations to an acceptable size, with a reasonably high value of $C_I$ needed in order for $\tilde{I}(0)$ to be not too large. From a clinical perspective, there may be restrictions on how this might be managed, so other strategies for immune cell injection could be considered based around optimal control strategies, such as bang-bang control [33].

Cancer cell density is an important metric in understanding the progression of a number of cancer-based diseases. In multiple myeloma, for example, mutated cancerous plasma cells can crowd the bone marrow, and significantly hinder the production of red blood cells, white blood cells and platelets. These cancerous cells can also escape the bone marrow, travel through the blood system and degrade many other body components. The agent-based model we have constructed can be considered as a very simple, but still realistic, proxy of the treatment of cancer crowding through immune cell injection in a multiple myeloma setting. Of course the processes underlying multiple myeloma in the bone marrow take place in an unstructured three-dimensional setting with very complex internal structures and interactions between a variety of proteins. As stated in Section 1, the ABM presented here makes some assumptions, namely that its two-dimensional domain, relatively simple interactions between cancer and immune cells, and movement of cancer cells are still able to appropriately capture this complexity. In particular, we do not intend to capture all protein interactions in detail, as there is little information about the interaction rates and introducing complexity without such information can have negative repercussions for equation learning. Regarding the learning of equations and their subsequent mathematical analysis, these assumptions can also be seen as a benefit, as they promote the learning of concise equations and allow drawing informative, high-level conclusions (such as the relation between $I_0$ and cancer cell density in Section 3.2).

Probably the most important abstraction made by the equation learning approach discussed here, is that no spatial information is considered. This is adequate to derive certain insights, but this generally depends on the research questions to be answered by the surrogate. In the future, equation learning might, possibly also considering space, be applied to more complex ABMs, but there is a tradeoff between capturing the full ABM dynamics and sparsity (analysability) of the reactions. Closely approximating the full dynamics of the ABM will result in ODEs with many terms and, depending on the variables of interest, additional ODEs that complicate the analysis. On the other hand, enforcing ODEs with fewer terms will reduce the fidelity of the surrogate to the ABM. By steering this tradeoff according to specific research questions, learning interpretable surrogate equations has the potential to lead to faster and new answers.

## Data availability statement

The code and data to reproduce all equation learning results is available at https://doi.org/10.5281/zenodo.15240282.

## Author contributions

KB: Conceptualization, Formal analysis, Investigation, Supervision, Visualization, Writing – original draft, Writing – review & editing. PB: Conceptualization, Formal analysis, Investigation, Supervision, Visualization, Writing – original draft, Writing – review & editing. JK: Data curation, Investigation, Methodology, Software, Visualization, Writing – original draft, Writing – review & editing. AU: Funding acquisition, Supervision, Writing – original draft, Writing – review & editing. HW: Data curation, Methodology, Software, Visualization, Writing – original draft.

## Funding

The author(s) declare that financial support was received for the research and/or publication of this article. This work is in part supported by the German Research Foundation (DFG) within the Collaborative Research Center (SFB) 1270/2 ELAINE 299150580 and by the ARC Centre of Excellence for Plant Success in Nature and Agriculture (CE200100015).

## Conflict of interest

The authors declare that the research was conducted in the absence of any commercial or financial relationships that could be construed as a potential conflict of interest.

## Generative AI statement

The author(s) declare that no Gen AI was used in the creation of this manuscript.

## Publisher's note